\documentclass{article}

\usepackage{amsmath, amssymb,cmll} 
\usepackage{multicol}
\usepackage{fourier}
\usepackage{bussproofs}
\usepackage{enumerate}
\usepackage{pgf}
\usepackage{tikz}

\usepackage{subcaption}
\usepackage{float}
\floatstyle{boxed}
\restylefloat{figure}
\usepackage{placeins}

\newcommand{\bc}{\begin{C}}
\newcommand{\ec}{\end{C}}
\newcommand{\be}{\begin{equation}}
\newcommand{\ee}{\end{equation}}

\newcommand{\nb}{\begin{Prop}}
\newcommand{\nbe}{\end{Prop}}
\newcommand{\bl}{\begin{LE}}
\newcommand{\el}{\end{LE}}

\newcommand{\bd}{\begin{Def}}
\newcommand{\ed}{\end{Def}}
\newcommand{\bt}{\begin{Th}}
\newcommand{\et}{\end{Th}}

\newtheorem{Th}{Theorem}[section]
\newtheorem{LE}[Th]{Lemma}
\newtheorem{C}[Th]{Corollary}

\newtheorem{Def}[Th]{Definition}
\newcommand{\bp}{\begin{Prop}}
\newcommand{\ep}{\end{Prop}}

\newtheorem{Prop}[Th]{Proposition}

\DeclareMathSymbol{\mlq}{\mathord}{operators}{``}
\DeclareMathSymbol{\mrq}{\mathord}{operators}{`'}

\renewenvironment{itemize}
  {\begin{olditemize}
     \setlength{\parskip}{0pt}}
  {\end{olditemize}}

\begin{document}
 \title{First order linear logic and tensor type calculus for categorial grammars
}
\author{Sergey Slavnov
\\  National Research University Higher School of Economics
\\ sslavnov@yandex.ru\\} \maketitle

\begin{abstract}
We study relationship between   first order multiplicative linear logic ({\bf MLL1}), which has been known to provide representations to different categorial grammars, and the recently introduced {\it extended tensor type calculus} ({\bf ETTC}). We identify a fragment of {\bf MLL1}, which seems sufficient for many grammar representations, and establish a correspondence between {\bf ETTC} and this fragment.  The system {\bf ETTC}, thus, can be seen as an alternative syntax and intrinsic deductive system together with a geometric representation for the latter. We also give a natural deduction formulation of {\bf ETTC}, which might be convenient.
\end{abstract}

\section{Introduction}
The best known examples of  categorial  grammars are  Lambek grammars, which are based on Lambek calculus ({\bf LC}) \cite{Lambek}, i.e. noncommutative intuitionistic  linear logic (for background on linear logic see \cite{Girard},\cite{Girard2}). These, however, have somewhat limited expressive power, and a lot of extensions/variations have been proposed, using discontinuous tuples of strings and $\lambda$-terms, commutative and non-commutative logical operations, modalities etc, let us mention displacement grammars \cite{Morrill_Displacement}, abstract categorial grammars  (also called $\lambda$-grammars and linear grammars) \cite{deGroote}, \cite{Muskens}, \cite{PollardMichalicek} and hybrid type logical categorial grammars \cite{KubotaLevine}.

It has been known for a while (starting from the seminal work \cite{MootPiazza}) that different grammatical formalisms, such as those just mentioned, can all be represented using simple and familiar commutative system of
{\it first order multiplicative intuitionistic linear logic} ({\bf MILL1})
  \cite{Moot_extended}, \cite{Moot_inadequacy}.  In fact, not
  the whole of {\bf MILL1} is used:
  it can be noted that translations of different categorial grammars usually fit into  some small fragment, which, therefore, can be given linguistic interpretation.  Unfortunately, we do not have any deductive system intrinsic to this fragment.
  Typically, when deriving the {\bf MILL1} translation of an {\bf LC} sequent  in sequent calculus or natural deduction, one might have to use at intermediate steps sequents and derivations which are not translations of anything and have no linguistic meaning at all (as it seems).

{\it Extended tensor type calculus} ({\bf ETTC}), a system extending propositional (classical) multiplicative linear logic ({\bf MLL}), recently proposed by the author \cite{Slavnov_tensor} (elaborating on \cite{Slavnov_cowordisms}), was directly designed for linguistic interpretation in terms of bipartite graphs whose  edges are labeled with words.
It has been shown that {\it tensor grammars} based on {\bf ETTC} include both Lambek grammars and abstract categorial grammars as conservative fragments, with representation very similar to that in {\bf MILL1}. However, unlike the case of {\bf MILL1}, in {\bf ETTC} derivations  each rule corresponds to a concrete operation on strings, so that  formal language generation is decomposed into elementary steps, which, moreover,  can be conveniently visualized in the pictorial setting of labeled graphs.

  In this work we identify a fragment (we call it {\it strictly balanced fragment}) of first order linear logic, which is sufficient for  linguistic constructions of \cite{Moot_extended}, \cite{Moot_inadequacy}, and show that there are mutually inverse translations to and from {\bf ETTC}. Thus {\bf ETTC} turns out to be an alternative syntax for the relevant fragment of {\bf MILL1}, equipped with an intrinsic deductive system and intuitive pictorial representation.

  We also introduce a natural deduction formulation for {\bf ETTC} in this work, which might be more convenient in some situations than the sequent calculus formulation of \cite{Slavnov_tensor}.

\section{Background: systems of linear logic}
 The language of {\it multiplicative intuitionistic logic} ({\bf MILL1}) is summarized in Figure \ref{MILL1 formulas}. We assume that we are given a  set $Pred$ of {\it predicate symbols} with assigned { arities}, a countable set $Var$ of {\it individual variables} and a set $Const$ of {\it constants}. The set of first order atomic formulas is denoted as $At$, and the set of all first order linear intuitionistic formulas is denoted as  $Fm$.
The binary connectives $\otimes$, $\multimap$ are called respectively {\it tensor} (also {\it times}) and {\it linear implication}. A {\it context} $\Gamma$ is a finite multiset of formulas;  as usual,  we denote formulas with Latin letters and contexts with Greek letters. The set of free variables in the context $\Gamma$ is denoted as $FV(\Gamma)$.  The sequent calculus for {\bf MILL1} is shown in Figure \ref{MILL11}.
\begin{figure}
\begin{subfigure}
{\textwidth}
\centering
$At=\{p(e_1,\ldots,e_n)|~e_1,\ldots,e_n\in Var\cup Const,~p\in Pred,~arity(p)=n\}$,
$Fm::=At|(Fm\multimap Fm)|(Fm\otimes Fm)|\forall xFm|\exists xFm,~x\in Var$.\\
\caption{{\bf MILL1} language}
\label{MILL1 formulas}
\end{subfigure}
\begin{subfigure}
{\textwidth}
\centering
\begin{multicols}{2}
$A\vdash A,
~A\in At~{\rm{(Id)}}
\quad
\cfrac{\Gamma\vdash A\quad A,\Theta\vdash C}{\Gamma,\Theta\vdash C}~{\rm{(Cut)}}
$\\
$
\cfrac{\Gamma,A,B\vdash C}{\Gamma, A\otimes B\vdash C}~{\rm{(\otimes L)}}
\quad\cfrac{\Gamma\vdash A\quad\Theta\vdash B}{\Gamma,\Theta\vdash A\otimes B}~{\rm{(\otimes R)}}$\\
$
\cfrac{\Gamma\vdash A\quad B,\Theta\vdash C}{\Gamma,A\multimap B,\Theta\vdash C}~{\rm{(\multimap L)}}
~\cfrac{\Gamma,A\vdash B}{\Gamma \vdash A\multimap B}~{\rm{(\multimap R)}}
$\\
$
\cfrac{\Gamma,A\vdash C}{\Gamma, \exists x A\vdash C},~x\not\in FV(\Gamma,C)~{\rm{(\exists L)}}
$\\
$
\quad\cfrac{\Gamma\vdash A}{\Gamma\vdash \forall x A},~x\not\in FV(\Gamma)~{\rm{(\forall R)}}
$\\
$
\quad\cfrac{\Gamma\vdash A[x:=t]}{\Gamma\vdash \exists xA}~{\rm{(\exists R)}}
~\cfrac{\Gamma,A[x:=t]\vdash C}{\Gamma, \forall x A\vdash C}{\rm{(\forall L)}}
$\\
\end{multicols}
\caption{{\bf MILL1} sequent calculus}
\label{MILL11}
\end{subfigure}
\begin{subfigure}
\textwidth
\centering
$Pred_-=\{\overline{p}|~p\in Pred_+\},\quad arity({\overline{p}})=arity({p}),\quad Pred=Pred_+\cup Pred_-$.\\
$Fm::=At|(Fm\otimes Fm)|(Fm\parr Fm)|\forall xFm|\exists xFm,~
x\in Var$\\
$\overline{p(e_1,\ldots,e_n)}=\overline{p}(e_n,\ldots,e_1),\quad
\overline{\overline{p(e_1,\ldots,e_n)}}=p(e_n,\ldots,e_1)\mbox{ for }P\in N_+,$
$\overline{A\otimes B}=\overline{B}\parr \overline{A},\quad \overline{A\parr B}=\overline{B}\otimes \overline{A},\quad
\overline{\forall x A}=\exists x(\overline{A}),\quad\overline{\exists x A}=\forall x(\overline{A})$.
\caption{\bf{MLL1} language}
\label{MLL1 formulas}
\end{subfigure}
\begin{subfigure}
\textwidth
\centering
\begin{multicols}{2}
${\vdash \overline{A},A},~A\in At~(\mbox{Id})\quad
\cfrac{\vdash \Gamma,A\quad\vdash
\overline{A},\Theta}{\vdash\Gamma,\Theta} ~(\mbox{Cut})$\\
$\cfrac{\vdash \Gamma,A,B}
{\vdash\Gamma,A\parr B}~
(\parr)\quad\cfrac{\vdash\Gamma, A \quad\vdash
B,\Theta}{\vdash\Gamma,A\otimes B,\Theta}~{ }
(\otimes)$\\
$\cfrac{\vdash\Gamma A}{\vdash \Gamma,\forall x A},~x\not\in FV(\Gamma)~(\forall)$\\
$\cfrac{\vdash\Gamma A[x:=e]}{\vdash \Gamma,\exists x A},~x\not\in FV(\Gamma)~(\exists)$.
\end{multicols}
\caption{{\bf MLL1} sequent calculus}
\label{MLL1}
\end{subfigure}
\begin{subfigure}
{\textwidth}
\centering
$A\multimap B=B\parr\overline{A}$.
$\quad
\begin{array}{rcl}
 A_1,\ldots,A_n\vdash_{\bf MILL1} B
 &\Longleftrightarrow&
 \vdash_{\bf MLL1} B,\overline{A_n},\ldots,\overline{A_1}.
 \end{array}$
\caption{Translating {\bf MILL1} to {\bf MILL1}}
\label{MILL2MLL}
\end{subfigure}
\begin{subfigure}
{\textwidth}
\centering
$Fm::=Prop|(Fm\backslash Fm)|(Fm/FM)| (Fm\bullet Fm)$\\
\begin{multicols}{2}
${A\vdash A}~(\mbox{Id})~\cfrac{\Gamma\vdash A\quad \Theta_1, A,\Theta_2\vdash B}{\Theta_1,\Gamma,\Theta_2\vdash B}~(\rm{Cut})$\\
$\cfrac{\Gamma, A\vdash B}{\Gamma \vdash B/A}~(/\rm{R})~
\cfrac{\Gamma\vdash A\quad \Theta_1, B,\Theta_2\vdash C}{\Theta_1,B/A,\Gamma,\Theta_2\vdash C}~(/\rm{L})$\\
$
\cfrac{\Theta_1\vdash A\quad \Theta_2\vdash B}{\Theta_1,\Theta_2\vdash A\bullet B}~(\bullet\rm{R})~ \cfrac{\Theta_1,A,B,\Theta_2\vdash C}{\Theta_1,A\bullet B,\Theta_2\vdash C}~(\bullet \rm{L})$
\\
$
\cfrac{A,\Gamma\vdash B}{\Gamma \vdash A\backslash B}~(\backslash \rm{R})~
\cfrac{\Gamma\vdash A\quad \Theta_1, B,\Theta_2\vdash C}{\Theta_1,\Gamma,A\backslash B ,\Theta_2\vdash C}~(\backslash\rm{L})$
\end{multicols}
\caption{{\bf LC} language and sequent rules}
\label{LC_rules}
\end{subfigure}
\begin{subfigure}
{\textwidth}
\centering
\begin{multicols}{2}
$||p||^{(l,r)}=p(l,r)\mbox{ for }p\in Prop$,\\
 $||A\bullet B||^{(l,r)}=\exists x||A||^{(l,x)}\otimes ||B||^{(x,r)}$,\\
$||B/ A||^{(l,r)}=\forall x||A||^{(r,x)}\multimap ||B||^{(l,x)}$,\\
$||A\backslash B||^{(l,r)}=\forall x||A||^{(x,l)}\multimap ||B||^{(x,r)}$.
\end{multicols}
$
\begin{array}{rcl}
A_1,\ldots, A_n\vdash_{\bf LC} B&
\Longleftrightarrow&||A_1||^{(c_0,c_1)},\ldots,||A_n||^{(c_{n-1},c_n)}\vdash_{\bf MILL1} ||B||^{(c_0,c_n)}.
\end{array}
$
\caption{Translating {\bf LC} to {\bf MILL1}}
\label{LC2MILL}
\end{subfigure}
\caption{Systems of linear logic}
\end{figure}
We will use  notation $\Gamma\vdash_{\bf MILL1}C$  to indicate that the sequent $\Gamma\vdash C$ is derivable in {\bf MILL1} and  a similar convention for other systems considered in this work.

The language of {\it classical first order multiplicative linear logic} ({\bf MLL1}) is summarized in Figure \ref{MLL1 formulas}.
We assume that we are given a set   $Pred_+$ of {\it positive predicate symbols} with assigned  arities, and sets $Var$ and $Const$ of  variables and constants. Then the set $Pred_-$ of {\it negative predicate symbols} and the set $Pred$ of all predicate symbols are defined.   Atomic formulas are defined same as for {\bf MILL1}.
Formulas are built from the binary connectives $\otimes$, $\parr$ and quantifiers,
the
connective $\parr$ is called  {\it cotensor} (also {\it par}).
{\it Linear negation} $\overline{(.)}$ is not a connective, but is {\it definable}
by induction on formula construction, as in Figure \ref{MLL1 formulas}.
Note  that, somewhat non-traditionally, we follow the convention that negation flips tensor/cotensor factors,  typical for {\it noncommutative} systems. This does not change the logic (the formulas $A\otimes B$ and $B\otimes A$ are provably equivalent), but is more consistent with our intended geometric interpretation.

Contexts are defined same as for {\bf MILL1}.
The {\it sequent calculus} for ${\bf MLL1}$  is shown in Figure \ref{MLL1}.
Translation in Figure \ref{MILL2MLL} identifies  {\bf MILL1}
as a {\it conservative fragment} of {\bf MLL1}.

The language and sequent calculus formulation of {\it Lambek calculus} ({\bf LC}) are summarised in Figure \ref{LC_rules}. Formulas are built from a set $Prop$  of {\it propositional symbols}, and contexts are {\it sequences}, rather than multisets, of formulas.

 Given two variables or constants $l, r$, the {\it first order translation} $||F||^{(l,r)}$
 of an {\bf LC} formula  $F$ parameterized by $l, r$
  is shown in Figure \ref{LC2MILL} ({\bf LC} propositional symbols are treated as binary predicate symbols). This embeds {\bf LC} into {\bf MILL1} as a conservative fragment \cite{MootPiazza}

\section{MILL1 grammars and strictly balanced fragment}
Translation from {\bf LC} suggests defining {\bf MILL1} grammars similar to Lambek grammars.

Let  a finite alphabet $T$ of terminal symbols be  given. Assume also that  our first order language contains all integer constants and two special constants $l,r$.

Let us say that a {\it  {\bf MILL1} lexical entry} is a pair $(w,A)$, where $w\in T^*$ is nonempty, and $A$ is a {\bf MILL1} formula with one occurrence of $l$ and one occurrence of $r$ and no other constants or free variables.
For the formula $A$ occurring in a simple lexical entry we will write
$A[x;y]=A[l:=x_1,r:=y_1]$
 (so that $A=A[l;r]$).

 A  {\bf MILL1} {\it grammar} $G$ is a pair $(Lex,S)$, where $Lex$ is a finite set of {\bf MILL1} lexical entries, and $S$ is a binary predicate symbol.
 The {\it language $L(G)$ generated by }$G$ or, simply, the {\it language of} $G$ is defined as
 $$
 L(G)=\{w_1\ldots w_n|~(w_1,A_1),\ldots, (w_m,A_n)\in Lex,~A_1[0;1],\ldots,A_n[{n-1};n]\vdash_{\bf MILL1} S(0,n)\}.
 $$
%
It seems clear that, under such a definition, Lambek grammars translate to {\bf MILL1} grammars generating the same language.

It has been shown \cite{Moot_extended}, \cite{Moot_inadequacy} that {\bf MILL1} allows representing more complex systems such as displacement calculus, abstract categorial grammars, hybrid type-logical grammars. This suggests that the above definition should be generalized to allow more complex lexical entries, corresponding to word tuples rather than just words.
On the other hand, it seems evident that in translations from grammatical formalisms we may restrict to smaller fragments of {\bf MILL1}.

%

\subsection{Balanced sequents}
We equip the first order language with an extra structure. We say that the language is {\it balanced} if each  $n$-ary predicate symbol $p$ is equipped with a {\it valency}, which is a pair $(k,m)$ of nonnegative integers with $k+m=n$. We indicate it in notation by writing corresponding atomic formulas as $p(x_1,\ldots,x_k;y_1,\ldots, y_m)$, with first $k$ entries separated by a semicolon. In this setting, we say that $x_1,\ldots,x_k$ are {\it left} occurrences,  or have {\it left polarity}, notation $sgn(x_i)=-1$, and $y_1,\ldots, y_m$ are  {\it right} occurrences,  $sgn(y_i)=+1$. We extend the notion of occurrence polarity  to compound formulas and sequents
 by induction.

In  the intuitionistic language the definition is as follows. For an occurrence $x$ in an immediate subformula $A$ of a compound formula $F$, the polarity $sgn_F(x)$ of $x$ in $F$ is defined by
$sgn_F(x)=            -sgn_A(x)$  if $F=A\multimap B$ and
            $sgn_F(x)=sgn_A(x)$ otherwise.
For  an occurrence $x$ in a formula $F$ of a sequent $\Gamma\vdash C$,  the polarity
$sgn_{\Gamma\vdash C}(x)$ of $x$ in the sequent is defined by
$sgn_{\Gamma\vdash C}(x)=
            -sgn_F(x)$, if $F\in\Gamma$, and
 $sgn_{\Gamma\vdash C}(x)=sgn_F(x)$ if $F=C$.

In the case of {\bf MLL1} we require that valencies of dual predicate symbols be consistent, so that $\overline{p(x_1,\ldots,x_k;y_1,\ldots,y_m)}=\overline{p}(y_m,\ldots,y_1;x_k,\ldots,x_1)$. Then the polarity of an occurrence in a formula or a sequent is just the same as its polarity in the corresponding atomic formula. Obviously, the embedding of {\bf MILL1} to {\bf MLL1} preserves polarities.

We say that a first order formula (context, sequent) is {\it balanced} if every quantifier binds exactly one left and one right variable occurrence. A balanced formula (context, sequent) is {\it strictly balanced} if, furthermore, it has at most one left and at most one right occurrence of any free variable or constant. It is immediate that if a strictly balanced  sequent is derivable (in {\bf MILL1} or {\bf MLL1} ) then every free variable or constant occurring in it has {\it exactly one} occurrence of each polarity.
We will say that  a {\bf MILL1} grammar is {\it balanced} if the formula in every lexical entry is strictly balanced, with constants $l$ and $r$ occurring with left and right polarity respectively.

It is immediate that if we treat {\bf LC} propositional symbols as predicate symbols of valency $(1,1)$, then the translation in Figure \ref{LC2MILL}  indeed uses only the strictly balanced fragment. Similar observations apply to translations in \cite{Moot_extended}, \cite{Moot_inadequacy}.

\subsection{Occurrence nets}
Below we are discussing derivations of balanced and strictly balanced sequents in {\bf MLL1}. The discussion can be adapted to the intuitionistic case using the standard embedding. In the following we will say simply {\it occurrence} meaning  occurrence of a  variable or a constant.
\begin{figure}
\centering
\begin{tikzpicture}
\begin{scope}[shift={(-3,0)}]
 \node at (0,0) {
$
\vdash a(e;t),\overline{a}(t;e)\otimes a(e;s),\overline{a}(s;e)$
};
 \draw[thick,-](-1.2,-0.2) to  [out=-30,in=180] (-.85,-.3) to  [out=0,in=-150](-0.5,-0.2);
 \draw[thick,-](1,-0.2) to  [out=-30,in=180] (1.35,-.3) to  [out=0,in=-150](1.7,-0.2);
 \draw[thick,-](-1.5,-0.2) to  [out=-30,in=180] (-.85,-.4) to  [out=0,in=-150](-.2,-0.2);
 \draw[thick,-](.7,-0.2) to  [out=-30,in=180] (1.35,-.4) to  [out=0,in=-150](2,-0.2);
 \end{scope}
 \node [above] at (0,0){$(\exists)$};
 \node at (0,-.1) {$\Longrightarrow$};
 \begin{scope}[shift={(3,0)}]

 \node at (0,0) {
$
\vdash a(e;t),\exists x(\overline{a}(t;x)\otimes a(x;s)),\overline{a}(s;e)$
};
 \draw[thick,-](-1.4,-0.2) to  [out=-30,in=180] (-.9,-.3) to  [out=0,in=-150](-0.4,-0.2);
 \draw[thick,-](1.2,-0.2) to  [out=-30,in=180] (1.6,-.3) to  [out=0,in=-150](2,-0.2);
 \draw[thick,-](-1.8,-0.2) to  [out=-30,in=180] (.25,-.5) to  [out=0,in=-150](2.3,-0.2);

 \end{scope}
\end{tikzpicture}
\caption{Occurrence nets example}
\label{occ_net}
\end{figure}

An {\it occurrence net} of a balanced {\bf MLL1} sequent $\vdash\Gamma$ is a
perfect matching $\sigma$ between left and right free occurrences in $\Gamma$, such that each pair ({\it link}) in $\sigma$ consists of occurrences of the same variable (constant). Note that for a strictly balanced sequent there is only one occurrence net possible.
Basically, occurrence nets are rudiments of proof-nets. To each cut-free derivation $\pi$ of  $\vdash\Gamma$ we assign by induction an occurrence net $\sigma(\pi)$.

 For an axiom $\vdash \overline{X}, X$, where
$X= {p}(e_1,\ldots,e_n)$,
  the net is defined by matching an occurrence  $e_i$ in $X$  with the occurrence $e_i$ in $\overline{X}=\overline{p}(e_n,\ldots,e_1)$. For $\pi$ obtained from a derivation $\pi'$ by the $(\parr)$ rule, we put $\sigma(\pi)=\sigma(\pi')$
  For $\pi$ obtained from derivations $\pi_1$, $\pi_2$ by the $(\otimes)$ rule, $\sigma(\pi)=\sigma(\pi_1)\cup\sigma(\pi_2)$.
If $\pi$ is obtained from some $\pi'$ by the $(\forall)$ rule introducing a formula $\forall x A$, then the variable $x$ has no free occurrences in the premise other than those in $A$, and, since the sequent is balanced, there is precisely one left and one right occurrence of $x$ in $A$, say, $x_l$ and $x_r$. It follows that $(x_l,x_r)\in\sigma(\pi')$ and we put $\sigma(\pi)=\sigma(\pi')\setminus\{(x_l,x_r)\}$. The rule erases the link $(x_l,x_r)$ from the occurrence net.

Finally, let $\pi$ be obtained from $\pi'$ by the $(\exists)$ rule applied to a  formula $A'=A[x:=e]$ and introducing the formula $\exists x A$. Since the sequent is balanced, there is exactly one left and one right occurrence of $x$ in $A$, and to them correspond a left and a right occurrence of $e$ in $A'$, say $e_l$ and $e_r$.
There are two cases.

If $(e_l, e_r)\in\sigma(\pi')$, put $\sigma(\pi)=\sigma(\pi')\setminus\{(e_l,e_r)\}$. The rule erases a link.
Otherwise let $e_l'$, $e_r'$ be such that $(e_l', e_r),(e_l, e_r')\in\sigma(\pi')$.
Put $\sigma(\pi)=\sigma(\pi')\setminus\{(e_l',e_r),(e_l,e_r')\}\cup\{(e_l',e_r')\}$. The rule glues two links  into one.
An  example for the latter case is shown in Figure \ref{occ_net}.

The following is easily proved by induction on derivation.
\bp\label{exists'}
Let $\pi$ be a cut-free derivation of a balanced sequent $\vdash\Gamma$, and assume that $(e_l,e_r)\in\sigma(\pi)$.
Let $e'$ be a fresh variable or constant and $\Gamma'=\Gamma[e_l,e_r:=e]$.

 Then $\vdash\Gamma'$ is derivable in {\bf MLL1} with a derivation $\pi'$ of the same size as $\pi$ and $\sigma(\pi')=\sigma(\pi)\setminus\{(e_l,e_r)\}\cup \{(e_l',e_r')\}$.
$\Box$
\ep

\subsection{Strictly balanced derivations}
Note that for all sequent rules except $(\exists)$, if the conclusion is strictly balanced so are the premises.
We will supply  an admissible rule $(\exists')$, alternative to $(\exists)$, which applies {\it to strictly balanced sequents only} and, therefore, satisfies the same property.

So assume that we have a derivable strictly balanced sequent $\vdash\Theta$, where $\Theta= \Gamma, A$. Assume that $A$ has a left  free occurrence $s_l$ and a right free occurrence $t_r$ of a variable (constant) $s,t$ respectively, with $s\not=t$. Let $e$ be a fresh variable or  constant.
Let $\Theta'=\Theta[s,t:=e]$.
By Proposition \ref{exists'} the sequent $\vdash\Theta'$ is derivable.

 Now, let us write  $\Theta'=\Gamma'A'$, where $A'$ is the image of $A$ in $\Theta'$.
Let $e_l$, $e_r$ be the occurrences of $e$ in $A'$ that replace
 $s_l$ and $t_r$ respectively. Then the strictly balanced
 sequent $\vdash\Theta''$, where
 $\Theta''=\Gamma',\exists x A'[e_l,e_r:=x]$,
  is derivable from $\vdash\Theta'$ by the $(\exists)$ rule.

We  say that $\vdash\Theta''$ is obtained from $\vdash\Theta$ {\it by the $(\exists')$ rule}. Note that, on the level of  occurrence nets, seen as bipartite graphs, the $(\exists')$ rule does the same gluing as $(\exists)$; only vertex labels are changed.

We will say that derivations of strictly balanced sequents using rules of {\bf MLL1} and the $(\exists')$ rule  and involving only strictly balanced sequents   are {\it strictly balanced derivations}.
\bl\label{strictly balanced fragment}
A  strictly balanced sequent $\vdash \Theta$   derivable in {\bf MLL1}  is derivable with a strictly balanced derivation.
\el
{\bf Proof} by induction on a cut-free derivation. If $\vdash\Theta$ is the conclusion of the $(\exists)$ rule applied to non-strictly balanced premise, then, using Proposition  \ref{exists'}, we replace the premise with a strictly balanced sequent and obtain $\vdash\Theta$ by the $(\exists')$ rule. $\Box$

Thus, adding the $(\exists')$ rule, we obtain a kind of intrinsic deductive system for the strictly balanced fragment. It seems though that the   usual syntax of first order sequent calculus is not very natural for such a system, and some other representation might be desirable.

\section{Tensor type calculus}\label{tensor types section}
\subsection{Tensor terms}\label{terms}
We assume that we are given   an infinite  set $\mathit{Ind}$ of  {\it indices}. They will be used in all syntactic objects (terms, types, typing judgements) that we consider.

Let $T$ be an alphabet of {\it terminal symbols}. The set
 $\widetilde{ TmEx}$ of {\it provisional tensor term expressions} is the free commutative monoid  generated by the set
\be\label{generators}
\{[w]^i_j|~i,j\in\mathit{Ind},w\in T^*\}\cup\{[w]|~w\in T^*\},
\ee
 and the set  $\widetilde{ Tm}$ of {\it provisional tensor terms} is the quotient of $\widetilde{ TmEx}$ by the relations
\be\label{tensor term relations}
[u]^j_i\cdot[v]^k_j=[uv]^k_i,\quad [w]^i_i=[w], \quad [a_1\ldots a_n]=[a_na_1\ldots a_{n-1}]\mbox{ for }a_1,\ldots,a_n\in T
\ee
(we write the monoid operation multiplicatively and the monoid unit as $1$).
 A {\it tensor term expression} is a provisional tensor term expression in which any index has at most one occurrence as an  upper one and at most one occurrence as a lower one. The set $Tm$ of {\it tensor terms} is the image of the set $TmEx$  of term expressions in $\widetilde{Tm}$ under the quotient map.
  (The adjective ``tensor'' will often be omitted in the following.)

Elements of generating set (\ref{generators}) are {\it elementary term expressions}.
Elementary  expressions  of the form $[w]^j_i$ with $i\not=j$ are {\it regular term expressions}.
A tensor term is {\it regular} if it is the image of the product of regular term expression. Otherwise the term is {\it singular}.

We think of regular terms  as bipartite graphs whose vertices are labeled with indices and edges with words, and direction of edges is from lower indices to upper ones. Singular terms such as $[w]^i_i=[w]$ correspond to closed loops (with no vertices) labeled with {\it cyclic} words (this explains the last relation in (\ref{tensor term relations})).
Thus, a regular term expression $[w]_i^j$, $i\not=j$, corresponds to a single edge from $i$ to $j$ labeled with $w$,
the product of two term expressions without common indices is  the disjoint union of the corresponding graphs, 
and a term expression with repeated indices corresponds to a graph obtained by gluing edges along matching vertices. The monoidal unit $1$ corresponds to the empty graph. As for singular terms, they arise when edges are glued cyclically.

 We say that an index occurring in a term expression $t$ is  {\it free} in $t$, if it occurs in $t$ once. Otherwise we say that the index is  {\it bound}.
 We say that a  term  expression is {\it normal} if it has no bound indices.
 The sets of free upper and lower indices of a term expression are invariant under the quotient map to $Tm$, so they are well-defined for terms as well.
We denote the set of  free upper, respectively lower, indices of a tensor term $t$   as $FSup(t)$, respectively $FSub(t)$.

The geometric representation makes  especially obvious   that
any  tensor term $t$ is the image of a  normal term expression and  can be written as the product $t=reg(t)\cdot t'$, where $reg(t)$ is regular and $t'$ is $1$ or the product of elementary singular expressions. We say that $reg(t)$ is  the {\it regular  part} of $t$.

Especially important are terms   of the form $\delta^j_i=[\epsilon]^j_i$, where $\epsilon$ denotes the empty word. We will also use the notation
$\delta^{i_1\ldots i_n}_{j_1\ldots j_n}=\delta^{i_1}_{j_1}\cdots\delta^{i_n}_{j_n}$.

We call terms of the above format {\it Kronecker deltas}. We will  adopt the  convention that capital Latin letters stand for sequences of indices and small Latin letters stand for individual indices. Typically, a Kreoneker delta will be written concisely as $\delta^I_J$.

Multiplication by  Kronecker deltas amounts to renaming indices.
If $t$ is a tensor term expression with
$i\in FSup(t)$, $j\in FSub(t)$, and
$i', j'\not\in \mathit{Ind}(t)$, then $\delta_{i}^{i'}\cdot t$ is the term $t$ with $i$ changed to $i'$, and
$\delta_{j'}^{j}\cdot t$ is $t$ with $j$ changed to $j'$.

There is also a ``divergent'' delta $\delta=\delta^i_i=[\epsilon]$ that corresponds to a closed loop with no label.

\subsection{Tensor types}
Our goal is to assign types to tensor terms.

The set $\widetilde{Tp}$ of {\it provisional tensor types} is built  according to the grammar in Figure \ref{ETTC_lang}.
We assume that we are given a set $Lit_+$ of  {\it positive atomic type symbols} or {\it positive literals}, and every  element $p\in Lit_+$ is assigned a {\it valency} $v(p)\in \mathbb{N}^2$.
Then the set $Lit_-$ of {\it negative atomic type symbols} and the set $Lit$ of all atomic type symbols are defined. The convention for negative atomic symbols is that $v(\overline{p})=(m,n)$ if  $v(p)=(n,m)$.
 Duality $\overline{(.)}$ is not a connective or operator but is definable. 

The symbols $\nabla,\Delta$ are {\it binding operators}.
These bind indices exactly in the same way as quantifiers bind  variables.
The operator $Q\in\{\nabla,\Delta\}$ in front of an expression $Q^i_jA$ has $A$ as its scope and binds all lower, respectively, upper occurrences of $i$, respectively, $j$ in $A$ that are not already bound by some other operator.
%
A {\it tensor type} is a provisional tensor type  in which any index has at most one free occurrence (no matter, upper or lower) and every binding operator binds exactly one lower and one upper index occurrence.
We will denote the set of upper, free upper,  lower, free lower indices occurring in a type $A$ as $Sup(A)$, $FSup(A)$, $Sub(A)$, $FSub(A)$ respectively. We say that tensor types are {\it $\alpha$-equivalent} if they differ by renaming bound indices in a usual way.
\begin{figure}
\begin{subfigure}{1\textwidth}
\centering
$Lit_-=\{\overline{p}|~p\in Lit_+\}$, $\quad\overline{\overline{p}}=p$ for $p\in Lit_+$, $\quad Lit=Lit_+\cup Lit_-$.\\
$\widetilde{At}=\{p_{j_1\ldots j_n}^{i_1\ldots i_m}|~p\in Lit,v(p)=(m,n),i_1,\ldots, i_m,j_1,\ldots,j_n\in\mathit{Ind}\}$.\\
$\widetilde{Tp}::=\widetilde{At}|(\widetilde{Tp}\otimes\widetilde{Tp})|(\widetilde{Tp}\parr\widetilde{Tp})|\nabla^i_j\widetilde{Tp}
|\Delta^i_j\widetilde{Tp},~
i,j\in\mathit{Ind},~i\not=j$.\\
$\overline{p^I_J}=\overline{p}^{\overline{J}}_{\overline{I}}\mbox{ for }p\in Lit$, where
 $\overline{I}=({i_n},\ldots, {i_1})$ for $I=(i_1,\ldots,i_n)$.
\\$ \overline{A\otimes B}=\overline{B}\wp\overline{A},\quad
\overline{A\wp B}=\overline{B}\otimes\overline{A},\quad
\overline{\nabla_i^jA}= \triangle^i_j\overline{A},\quad
\overline{\triangle_i^jA}= \nabla^i_j\overline{A}.
$
\caption{Language}
\label{ETTC_lang}
\end{subfigure}
\begin{subfigure}{1\textwidth}
\centering
$
\begin{array}{rl}
\delta^{I'\overline{J}}_{\overline{I}J'}\vdash {p}^{I}_{J},\overline{p}_{I'}^{J'},~p\in Lit~({\rm{Id}})
&
\cfrac{t\vdash \Gamma,A\quad s\vdash
\overline{A},\Theta}{ts\vdash\Gamma,\Theta} ~({\rm{Cut}})\\
\cfrac{t\vdash
\Gamma,A,B}{t\vdash
\Gamma,A\parr B}
~
(\parr)~
\cfrac{t\vdash\Gamma, A \quad s\vdash
B,\Theta}{ts\vdash\Gamma,A\otimes B,\Theta}~(\otimes)
&
\cfrac{
    \delta_\beta^\alpha\cdot t\vdash\Gamma,A
    }
    {
    t\vdash\Gamma,\nabla^\alpha_\beta A
    }~(\nabla)~
\cfrac{
    t\vdash\Gamma,A
    }
    {
    \delta^\beta_\alpha\cdot t\vdash\Gamma,\triangle_\beta^\alpha A
    }~(\triangle).
\end{array}
$
\caption{{\bf ETTC} typing rules}
\label{ETTC}
\end{subfigure}
\begin{subfigure}{1\textwidth}
\centering
         \begin{tikzpicture}[xscale=2,yscale=2]
    \begin{scope}[shift={(.5,0)}]
        \draw[thick,->](1.25,0) to  [out=-90,in=0] (.625,-.5) to  [out=180,in=-90](0,0);
        \node at(.625,-.4){xy};
        \node at(.45,-.15){ba};
        \draw[thick,->](1,0) to  [out=-90,in=0] (.75,-.25) to  [out=180,in=-90](0.5,0);
        \draw[thick,<-](0.75,0) to  [out=-90,in=0] (.5,-.25) to  [out=180,in=-90](.25,0);

        \draw [fill] (.25,0) circle [radius=0.05];
        \draw [fill] (1,0) circle [radius=0.05];
        \draw [fill] (1.25,0) circle [radius=0.05];

        \node [above]at(0,0){$i$};
        \node [above]at(0.25,0){$j$};
        \node [above]at(0.5,0){$k$};
        \node [above]at(0.75,0){$l$};
        \node [above]at(1,0){$r$};
        \node [above]at(1.25,0){$s$};
        \end{scope}
        \node at (2.5,-.15){$\Leftrightarrow$};
         \node[right] at (3.1,-.15) {$[xy]_s^i\cdot [ba]_j^l\cdot[\delta]_r^k\vdash a_i\otimes b^j_{kl}, c^{rs}$};

        \end{tikzpicture}
        \caption{Picture for a typing judgement}
        \label{Typing_judgement_geometrically}
        \end{subfigure}
    \begin{subfigure}{.5\textwidth}
        \centering
         \begin{tikzpicture}[xscale=.75,yscale=.75]

         \begin{scope}[xscale=-1.75,yscale=1.5]
        \begin{scope}[shift={(2.6,1)}]

        \begin{scope}[shift={(-2.5,0)}]
          \draw[thick,<-](0.8,0) to  [out=-90,in=180] (1.3,-.25) to  [out=0,in=-90](1.8,0);
          \draw[thick,<-](.5,0) to  [out=-90,in=180] (1.3,-.5) to  [out=0,in=-90](2.1,0);
          \draw[thick,->](0.3,0) to  [out=-90,in=180] (1.3,-.75) to  [out=0,in=-90](2.3,0);
          \draw[thick,->](0,0) to  [out=-90,in=180] (1.3,-1) to  [out=0,in=-90](2.6,0);

                \draw [fill] (0,0) circle [radius=0.05];
                \draw [fill] (0.15,0) circle [radius=0.01];
                \draw [fill] (0.2,0) circle [radius=0.01];
                \draw [fill] (0.1,0) circle [radius=0.01];
                        \draw [fill] (0.3,0) circle [radius=0.05];
                \draw[dashed,-](0,.05)--(0,.15)--(0.3,.15)--(0.3,.05);
                \node at (0.15,.15) [above] {${I}$};
                \node at (-.9,-.4)
                         {($\delta_{I'\overline{J}}^{\overline{I}J'}\vdash \overline{p}^{I'}_{J'},p_{I}^{J}$)};

                \draw[dashed,-](0,.15)--(0,.55)--(0.8,.55)--(0.8,.15);
                \node at (0.4,.55) [above] {$\overline{p}$};

                \begin{scope}[shift={(.5,0)}]
                \draw [fill] (0,0) circle [radius=0.01];
                \draw [fill] (0.15,0) circle [radius=0.01];
                \draw [fill] (0.2,0) circle [radius=0.01];
                \draw [fill] (0.1,0) circle [radius=0.01];
                \draw [fill] (0.3,0) circle [radius=0.01];
                \draw[dashed,-](0,.05)--(0,.15)--(0.3,.15)--(0.3,.05);
                \node at (0.15,.15) [above] {${J}$};
                \end{scope}

            \begin{scope}[shift={(1.8,0)}]
            \draw [fill] (0,0) circle [radius=0.05];
                \draw [fill] (0.15,0) circle [radius=0.01];
                \draw [fill] (0.2,0) circle [radius=0.01];
                \draw [fill] (0.1,0) circle [radius=0.01];
                        \draw [fill] (0.3,0) circle [radius=0.05];
                \draw[dashed,-](0,.05)--(0,.15)--(0.3,.15)--(0.3,.05);
                \node at (0.15,.15) [above] {$J'$};

                \draw[dashed,-](0,.15)--(0,.55)--(0.3,.55)--(0.5,.55);
                \node at (0.4,.55) [above] {${p}$};

                \begin{scope}[shift={(.5,0)}]
                \draw [fill] (0,0) circle [radius=0.01];
                \draw [fill] (0.15,0) circle [radius=0.01];
                \draw [fill] (0.2,0) circle [radius=0.01];
                \draw [fill] (0.1,0) circle [radius=0.01];
                \draw [fill] (0.3,0) circle [radius=0.01];
                \draw[dashed,-](0,.02)--(0,.15)--(0.3,.15)--(0.3,.05);
                \node at (0.15,.15) [above] {$I'$};

                \draw[dashed,-](0,.55)--(0.3,.55)--(0.3,.15);
                \end{scope}
            \end{scope}
          \end{scope}

          \end{scope}
          \end{scope}
\end{tikzpicture}
\caption{Picture for an axiom}
\label{axiom geometrically}
\end{subfigure}
\begin{subfigure}{.5\textwidth}
        \centering
         \begin{tikzpicture}[xscale=.6,yscale=.6]
\begin{scope}[shift={(10.5,0)}]
 \begin{scope}[shift={(1,0)}]

 \draw[thick](-5,0)rectangle(-3.5,1);\node at(-4.25,.5){$t$};
  \draw[thick](-3,0)rectangle(-1.5,1);\node at(-2.25,.5){$s$};
 \draw[thick,-](-4.75,1)--(-4.75,1.5);

 \node[above] at(-4.75,1.5) {$\Gamma$};
 \draw[thick,-](-3.75,1)--(-3.75,1.5);
 \node[above] at(-3.75,1.5) {$a$};

 \draw[thick,-](-1.75,1)--(-1.75,1.5);

 \node[above] at(-1.75,1.5) {$\Delta$};
 \draw[thick,-](-2.75,1)--(-2.75,1.5);
 \node[above] at(-2.75,1.5) {$\overline{a}$};
 \node at (-.75,.5) {$\Rightarrow$};
 \end{scope}

\begin{scope}[shift={(-4,0)}]
     \draw[thick](5,0)rectangle(6.5,1);\node at(5.75,.5){$t$};
 \draw[thick,-](5.25,1)--(5.25,1.5);

 \node[above] at(5.25,1.5) {$\Gamma$};
 \draw[thick,-](5.75,1)to[out=90,in=180](6.25,1.5)
 to[out=0,in=90](7.75,1);

         \draw[thick](7.,0)rectangle(8.5,1);\node at(8,.5){$s$};

 \draw[thick,-](8.25,1)--(8.25,1.5);

 \node[above] at(8.25,1.5) {$\Delta$};
 \node at (9.25,.5) {$(\mbox{cut})$};
 \end{scope}
 \end{scope}

        \end{tikzpicture}
        \caption{Picture for a cut}
        \label{cut_geometrically}
        \end{subfigure}
\begin{subfigure}{.5\textwidth}
        \centering
\begin{tikzpicture}[xscale=1.25,yscale=1.]
         \draw[draw=black](1.2,-.2)rectangle(-.5,-.5);\node at(.35,-.35){$t$};

         \begin{scope}[shift={(.1,0)}]
            \begin{scope}[shift={(-.3,.2)}]
                \draw [fill] (-.1,0) circle [radius=0.01];
                \draw [fill] (-0.05,0) circle [radius=0.01];
                \draw [fill] (0,0) circle [radius=0.01];
                \draw [fill] (0.05,0) circle [radius=0.01];
                        \draw [fill] (0.1,0) circle [radius=0.01];
                \draw[thick,dashed, -](-.2,0)--(-.2,.35) -- (.2,.35)-- (.2,0);
                        \node at (0,.3) [above] {$\Gamma$};
            \end{scope}

                \draw[thick, -](-.5,.2)--(-.5,-.2);
                \draw[thick, -](-.1,.2)--(-.1,-.2);

         \end{scope}

                \begin{scope}[shift={(.5,.2)}]

                \draw[thick,-](-.25,0)--(-.25,-.4);
                \draw[thick,-](.55,0)--(.55,-.4);

                \draw [fill] (-.1,0) circle [radius=0.01];
                \draw [fill] (-.15,0) circle [radius=0.01];
                \draw [fill] (0.15,0) circle [radius=0.01];
                \draw [fill] (0.2,0) circle [radius=0.01];
                \draw [fill] (0.1,0) circle [radius=0.01];
                \draw [fill] (.4,0) circle [radius=0.01];
                \draw [fill] (.45,0) circle [radius=0.01];

                \draw[thick,dashed, -](-.25,0)--(-.25,.35) -- (.55,.35)-- (.55,0);
                        \node at (.15,.3) [above] {$a$};

                \begin{scope}[shift={(0,-.05)}]
                    \draw [fill] (0.3,0) circle [radius=0.01];
                    \draw [fill] (0,0) circle [radius=0.05];
                    \node at (0,.25)  {$\beta$};
                    \node at (0.3,.0) [above] {$\alpha$};
                    \draw[thick,->](0,0) to  [out=-90,in=180] (.15,-.25) to  [out=0,in=-90](0.3,0);
                \end{scope}

                \end{scope}

\node at(1.6,-.35){$\Rightarrow$};

            \begin{scope}[shift={(2.5,0)}]
         \draw[draw=black](1.2,-.2)rectangle(-.5,-.5);\node at(.35,-.35){$t$};

         \begin{scope}[shift={(.1,0)}]
            \begin{scope}[shift={(-.3,.2)}]
                \draw [fill] (-.1,0) circle [radius=0.01];
                \draw [fill] (-0.05,0) circle [radius=0.01];
                \draw [fill] (0,0) circle [radius=0.01];
                \draw [fill] (0.05,0) circle [radius=0.01];
                        \draw [fill] (0.1,0) circle [radius=0.01];
                \draw[thick,dashed, -](-.2,0)--(-.2,.35) -- (.2,.35)-- (.2,0);
                        \node at (0,.3) [above] {$\Gamma$};
            \end{scope}

                \draw[thick, -](-.5,.2)--(-.5,-.2);
                \draw[thick, -](-.1,.2)--(-.1,-.2);

         \end{scope}

                \begin{scope}[shift={(.5,.2)}]

                \draw[thick,-](-.25,0)--(-.25,-.4);
                \draw[thick,-](.55,0)--(.55,-.4);

                \draw [fill] (-.1,0) circle [radius=0.01];
                \draw [fill] (-.15,0) circle [radius=0.01];
                \draw [fill] (0.15,0) circle [radius=0.01];
                \draw [fill] (0.2,0) circle [radius=0.01];
                \draw [fill] (0.1,0) circle [radius=0.01];
                \draw [fill] (.4,0) circle [radius=0.01];
                \draw [fill] (.45,0) circle [radius=0.01];

                \draw[thick,dashed, -](-.25,0)--(-.25,.35) -- (.55,.35)-- (.55,0);
                        \node at (.15,.3) [above] {$\nabla_\beta^\alpha a$};

                \begin{scope}[shift={(0,-.05)}]
                \end{scope}

                \end{scope}

                \end{scope}

        \end{tikzpicture}
        \caption{Picture for a $(\nabla)$ rule}
        \label{nabla_geometrically}
        \end{subfigure}
\begin{subfigure}{.5\textwidth}
        \centering
         \begin{tikzpicture}[xscale=1.25,yscale=1.]
         \draw[draw=black](1.2,-.2)rectangle(-.5,-.5);\node at(.35,-.35){$t$};

         \begin{scope}[shift={(.1,0)}]
            \begin{scope}[shift={(-.3,.2)}]
                \draw [fill] (-.1,0) circle [radius=0.01];
                \draw [fill] (-0.05,0) circle [radius=0.01];
                \draw [fill] (0,0) circle [radius=0.01];
                \draw [fill] (0.05,0) circle [radius=0.01];
                        \draw [fill] (0.1,0) circle [radius=0.01];
                \draw[thick,dashed, -](-.2,0)--(-.2,.35) -- (.2,.35)-- (.2,0);
                        \node at (0,.3) [above] {$\Gamma$};
            \end{scope}

                \draw[thick, -](-.5,.2)--(-.5,-.2);
                \draw[thick, -](-.1,.2)--(-.1,-.2);

         \end{scope}

                \begin{scope}[shift={(.5,.2)}]

                \draw[thick,-](-.25,0)--(-.25,-.4);
                \draw[thick,-](.55,0)--(.55,-.4);

                \draw[thick,->](0,-.05)--(0,-.4);
                \draw[thick,<-](.3,-.05)--(.3,-.4);

                \draw [fill] (-.1,0) circle [radius=0.01];
                \draw [fill] (-.15,0) circle [radius=0.01];
                \draw [fill] (0.15,0) circle [radius=0.01];
                \draw [fill] (0.2,0) circle [radius=0.01];
                \draw [fill] (0.1,0) circle [radius=0.01];
                \draw [fill] (.4,0) circle [radius=0.01];
                \draw [fill] (.45,0) circle [radius=0.01];

                \draw[thick,dashed, -](-.25,0)--(-.25,.35) -- (.55,.35)-- (.55,0);
                        \node at (.15,.3) [above] {$a$};

                \begin{scope}[shift={(0,-.05)}]
                    \draw [fill] (0.3,0) circle [radius=0.01];
                    \draw [fill] (0,0) circle [radius=0.05];
                    \node at (0,0.2)  {$\beta$};
                    \node at (0.3,.0) [above] {$\alpha$};
                \end{scope}

         \end{scope}

\node at(1.6,-.35){$\Rightarrow$};

            \begin{scope}[shift={(2.5,0)}]
         \draw[draw=black](1.2,-.2)rectangle(-.5,-.5);\node at(.35,-.35){$t$};

         \begin{scope}[shift={(.1,0)}]
            \begin{scope}[shift={(-.3,.2)}]
                \draw [fill] (-.1,0) circle [radius=0.01];
                \draw [fill] (-0.05,0) circle [radius=0.01];
                \draw [fill] (0,0) circle [radius=0.01];
                \draw [fill] (0.05,0) circle [radius=0.01];
                        \draw [fill] (0.1,0) circle [radius=0.01];
                \draw[thick,dashed, -](-.2,0)--(-.2,.25) -- (.2,.25)-- (.2,0);
                        \node at (0,.25) [above] {$\Gamma$};
            \end{scope}

                \draw[thick, -](-.5,.2)--(-.5,-.2);
                \draw[thick, -](-.1,.2)--(-.1,-.2);

         \end{scope}

                \begin{scope}[shift={(.5,.2)}]

                \draw[thick,-](-.25,0)--(-.25,-.4);
                \draw[thick,-](.55,0)--(.55,-.4);

                \draw[thick,-](0,-.05)--(0,-.4);
                \draw[thick,-](.3,-.05)--(.3,-.4);

                \draw [fill] (-.1,0) circle [radius=0.01];
                \draw [fill] (-.15,0) circle [radius=0.01];
                \draw [fill] (0.15,0) circle [radius=0.01];
                \draw [fill] (0.2,0) circle [radius=0.01];
                \draw [fill] (0.1,0) circle [radius=0.01];
                \draw [fill] (.4,0) circle [radius=0.01];
                \draw [fill] (.45,0) circle [radius=0.01];

                \draw[thick,dashed, -](-.25,0)--(-.25,.25) -- (.55,.25)-- (.55,0);
                        \node at (.15,.25) [above] {$\triangle_\beta^\alpha a$};

                \begin{scope}[shift={(0,-.05)}]
                    \draw[thick,-](0,0) to  [out=90,in=180] (.15,.2) to  [out=0,in=90](0.3,0);
                \end{scope}

         \end{scope}
                \end{scope}

        \end{tikzpicture}
        \caption{Picture for a $(\triangle)$ rule}
        \label{delta_geometrically}
        \end{subfigure}
\begin{subfigure}{.3\textwidth}
\centering
\begin{tikzpicture}
\begin{scope}[xscale=1]
\draw[thick,<-](0,0) to  [out=-90,in=180] (.75,-.6) to  [out=0,in=-90](1.5,0);
\draw[thick,->](-.4,0) to  [out=-90,in=180] (.75,-1) to  [out=0,in=-90](1.9,0);
        \draw [fill] (-.4,0) circle [radius=0.05];
        \draw [fill] (1.5,0) circle [radius=0.05];
        \node [above] at (-0.2,0) {$b$};
        \node [above] at (1.75,0) {$\overline{a}$};

        \node  at (0.75,-.8) {string1};
        \node [above] at (0.75,-.6) {string2};
\end{scope}
\end{tikzpicture}
\caption{Picture for $a\multimap b$}
\label{implication_general_form}
\end{subfigure}
\begin{subfigure}{.3\textwidth}
\centering
\begin{tikzpicture}
\begin{scope}[xscale=1]
\draw[thick,<-](0,0) to  [out=-90,in=180] (.75,-.6) to  [out=0,in=-90](1.5,0);
\draw[thick,->](-.4,0) to  [out=-90,in=180] (.75,-1) to  [out=0,in=-90](1.9,0);
        \draw [fill] (-.4,0) circle [radius=0.05];
        \draw [fill] (1.5,0) circle [radius=0.05];
        \node [above] at (-0.2,0) {$b$};
        \node [above] at (1.75,0) {$\overline{a}$};

        \node  at (0.75,-.8) {string};

\end{scope}
\end{tikzpicture}
\caption{$b/a$ inside $a\multimap b$}
\label{implication_/}
\end{subfigure}
\begin{subfigure}{.3\textwidth}
\centering
\begin{tikzpicture}
\begin{scope}[xscale=1]
\draw[thick,<-](0,0) to  [out=-90,in=180] (.75,-.6) to  [out=0,in=-90](1.5,0);
\draw[thick,->](-.4,0) to  [out=-90,in=180] (.75,-1) to  [out=0,in=-90](1.9,0);
        \draw [fill] (-.4,0) circle [radius=0.05];
        \draw [fill] (1.5,0) circle [radius=0.05];
        \node [above] at (-0.2,0) {$b$};
        \node [above] at (1.75,0) {$\overline{a}$};

        \node [above] at (0.75,-.6) {string
        };
\end{scope}

\end{tikzpicture}
\caption{$a\backslash b$ inside $a\multimap b$}
\label{implication_setmnus}
\end{subfigure}
\begin{subfigure}{1\textwidth}
\centering
 $a^I_J\multimap b^K_L=b^K_L\parr\overline{a}^I_J,\quad(b/a)^i_j=\nabla^\alpha_\beta(b^i_\alpha\wp \overline{a}_j^\beta ),\quad
(a\backslash b)^i_j=\nabla^\alpha_\beta(\overline{a}_\alpha^i\wp b_j^\beta ).$
\caption{Encoding different implications in {\bf ETTC}}
\label{translating lambek types}
\end{subfigure}
        \caption{{\bf ETTC}}
        \label{rules geometrically}
   \end{figure}

%

We also define  tensor type symbols as, basically, $\alpha$-equivalence classes of tensor types with all free indices erased. That is,
{\it tensor type symbol} is an equivalence class for the smallest equivalence relation on tensor types that identifies  any two $\alpha$-equivalent types  and any two types obtained from each other by renaming a free index.
{\it Valency} $v(a)$ of a type symbol $a$ is a pair $(n,m)$, where $n$ and $m$ are the numbers of, respectively, free upper and free lower indices in corresponding types.
%
%

Usually we will denote types with capital letters and type symbols with small letters.  A tensor type, up to $\alpha$-equivalence, can be recovered from its symbol by specifying the sequences of its upper and lower free indices.
Accordingly, we will write $A=a^I_J$ to indicate that $A$ is a type whose symbol is $a$, and sequences of free upper and lower indices are $I$ and $J$ respectively. Sometimes we will have enumerations of type symbols, such as $a_{(1)},\ldots,a_{(n)}$. In such  cases we will put brackets around subscripts in order  to avoid confusion with tensor type indices.


A {\it tensor  type context}   is a finite set of tensor types  whose elements have no common free indices.
We extend the notation for  sets of upper, free upper, lower and free lower indices from types to type contexts in the obvious way and write
$Sup(\Gamma)=\bigcup\limits_{A\in\Gamma}Sup(A)$ etc.

\subsection{Typing judgements and rules}
A {\it  tensor typing judgement}  is a pair $(t,\Gamma)$, written as $t\vdash \Gamma$, where
 $\Gamma$ is a tensor  type context,
  and $t$ is a tensor term, such that $FSup(t)=FSub(\Gamma)$, $FSub(t)=FSup(\Gamma)$.


Typing judgements are derived using the  rules of {\it extended tensor type calculus} ({\bf ETTC}) \cite{Slavnov_tensor} in Figure \ref{ETTC}. (The title ``extended'' refers in \cite{Slavnov_tensor} to usage of binding operators, which extend plain types of {\bf MLL}.)

  It is implicit in the rules  that all typing judgements are well defined, i.e. there are no index collisions.
In the $(\nabla)$ and $(\Delta)$ rules it is required that $\alpha\in FSub(A)$, $\beta\in FSup(A)$.
Note that the term $t$ in the premise of the $(\triangle)$ rule must have { free} occurrences of $\alpha$ and $\beta$ (as an upper and a lower index respectively).
 In the conclusion,  the term $\delta^\beta_\alpha\cdot t$ has these occurrences {\it bound}. Thus free indices to the left and to the right of the turnstile do match, and the typing judgement  is well defined.

It is not hard to see that {\bf ETTC} is cut-free \cite{Slavnov_tensor}. Note also that the system does not use any terminal alphabet. Terminal symbols will appear in {\it nonlogical axioms}, i.e. lexical entries of formal grammars.

 Let us define
 {\it$\alpha$-equivalence of tensor typing judgements} as the smallest equivalence relation identifying typing judgements that can be obtained from each other  by renaming bound indices in types or, for a free index, by replacing  both its occurrences  to the right and to the left of the turnstile with a fresh one.
 Then, obviously, derivable typing judgements are closed under $\alpha$-equivalence.
  $\alpha$-Equivalent typing judgements are essentially ``the same''; nothing is lost even if we identify them as in \cite{Slavnov_tensor}.

  \subsubsection{Geometric representation and meaning}
  While tensor terms have a natural interpretation as edge-labeled graphs, tensor typing judgements  suggest some particular pictorial representation for such graphs. A type to the right of the turnstile can be seen as inducing an ordering on the set of {\it free} indices, for example, from left to right, from   top to bottom, and this can be seen as an ordering of vertices in a picture. Thus, a typing judgement  $t\vdash F$ can be depicted as the graph representing the term $t$ with vertices aligned, say, horizontally according to the ordering induced by $F$. (Bound indices of $F$ are not in the picture.)
An example of a concrete  typing judgement representation is given  in Figure \ref{Typing_judgement_geometrically}.
Note that such a representation identifies $\alpha$-equivalent judgements, and the indices usually become redundant, at least, in the absence of binding operators.

   Some schematic pictures corresponding to rules of {\bf ETTC} are shown in Figure \ref{rules geometrically}.
   The $(\triangle)$ rule simply glues together  two bound indices/vertices. On the other hand, the $(\nabla)$ rule is applicable only in the case when the corresponding indices/vertices are connected with an edge carrying no label. Then this edge (together with its endpoints) is erased from the picture completely. The information about the erased edge is stored in the introduced type.

      Thus, terms of type $\nabla^\alpha_\beta A$ encode the {\it subtype} of $A$ consisting of terms/graphs with that specific form: vertices corresponding to $\alpha$ and $\beta$ are connected with an edge, and the connecting edge carries no label.
      It can be observed that
      we have the admissible rule
      $
      \cfrac{t\vdash \Gamma,\nabla^\alpha_\beta A}{\delta^\alpha_\beta \cdot t\vdash \Gamma,A}~(\nabla{\rm{E}})$, so that ``decoding'' from $\nabla^\alpha_\beta A$ to $A$ is always possible.

Let $a$, $b$ be type symbols of valency $(1,1)$, so that  regular elements of the corresponding types are strings.
The implication type symbol $a\multimap b=b\wp\overline{a}$ has valency $(2,2)$, and elements
of the corresponding type can be interpreted as pairs of strings, as Figure \ref{implication_general_form} suggests.
Now, there are two subtypes consisting of elements of the forms respectively
$
[u]_l^i\cdot\delta_k^j$  and $[u]_k^j\cdot\delta_l^i$
corresponding to Figures \ref{implication_/}, \ref{implication_setmnus}.
 It is easily computed that elements of the first form act (by means of the Cut rule) on elements of $a_k^l$ by multiplication (concatenation) on the left, and elements of the second form, by multiplication on the right.
 The two formats  identify two {\it subtypes} of the implicational tensor type that correspond to two implicational types of Lambek calculus. Which suggests that the implicational types of Lambek calculus should be translated to {\bf ETTC} as in Figure \ref{translating lambek types}
 (compare with Figure \ref{LC2MILL}).

\subsection{Relation with first order logic}
Given a balanced first order language, we identify predicate symbols with atomic type symbols   of the same valencies.
For  an {\bf MLL1} strictly balanced context $\Gamma$, let $X$ be the set variables and constants occurring in $\Gamma$ (not necessarily freely) and choose non-intersecting subsets $I,J\subset\mathit{Ind}$ together with a pair of bijections $\pi:X\to I$, $\rho:X\to J$. Translation of subformulas  of $\Gamma$ to  tensor types  is
given by
$$
||a(x_1,\ldots,x_n;y_1,\ldots,y_m)||^{\rho}_\pi=a^{\rho(x_1)\ldots \rho(x_n)}_{\pi(y_1)\ldots \pi(y_m)}\mbox{ for } a\in N,$$
$$||\forall x A||^{\rho}_\pi=
\nabla^{\pi(x)}_{\rho(x)}||A||^{\rho}_\pi,
\quad||\exists x A||^{\rho}_\pi=
\Delta^{\pi(x)}_{\rho(x)}||A||^{\rho}_\pi$$
 (with binary connectives obviously translating to themselves).
\bl\label{FO2ETTC}
There is a translation assigning to any strictly balanced derivation of $\vdash\Gamma$
a derivable tensor typing judgement $\tau\vdash||\Gamma||^{\rho}_\pi$, where the regularization $reg(\tau)$ is, independent of a particular derivation, the product of all Kronecker deltas $\delta^{\pi(x)}_{\rho(x)}$, where $x$ ranges over all free occurrences in $\Gamma$. Conversely, any derivable tensor typing judgement is obtained as a translation of a strictly balanced {\bf MLL1} derivation.
\el
{\bf Proof} The translation is by induction on strictly balanced derivation. The regular part $reg(\tau)$  encodes the unique occurrence net of $\Gamma$. Axioms and propositional rules translate to themselves. The $(\forall)$ rule  translates to the $(\nabla)$ rule,
while the $(\exists)$ and  $(\exists')$ rules  translate as the $(\Delta)$ rule.
 An interesting case is that of the $(\exists)$ rule. When
applied to a strictly balanced sequent, it  erases a link from the occurrence net, just as the $(\forall)$ rule.
As for
the $(\Delta)$ rule in this case, it glues together the corresponding link endpoints, producing a closed loop, divergent $\delta$. The link is removed from the regular part, but not from the term entirely.
The second claim of the lemma is proven by induction on {\bf ETTC} derivations. $\Box$

We observe that, in general, the tensor translation does indeed  depend on the derivation, only the regular part of the term is completely determined by the sequent.

%

\section{Tensor natural deduction and grammars}
In order to give a natural deduction formulation for {\bf ETTC} we need to allow variables standing for tensor terms. Thus we introduce a countable set of {\it tensor variable symbols} of different valencies (where valency, as usual, is a pair of nonnegative integers) and the convention that if $x$ is a variable symbol of valency $(m,n)$ and $i_1,\ldots,i_m$, $j_1,\ldots,j_m$ are pairwise distinct indices, then  $x^{i_1\ldots i_n}_{j_1\ldots j_m}$ is a {\it tensor variable}. Similarly to types and type symbols,
 we will use small letters for variable symbols and capital letters for variables; typically, we will write $X=x^I_J$.

{\it Natural deduction tensor term expressions and terms} are defined from provisional term expressions exactly as in Section \ref{terms}, with the  modification that we add
  to generators (\ref{generators}) of the provisional monoid $\widetilde{TmEx}$ all tensor variables, and  to defining relations (\ref{tensor term relations}) all possible equations
 $$\delta^i_jx_{IiJ}^K=x_{IjJ}^K,\quad \delta^i_jx^{IjJ}_K=x^{IiJ}_K.$$
In the following we abbreviate the title ``natural deduction'' as ``n.d.''.

\begin{figure}[t]
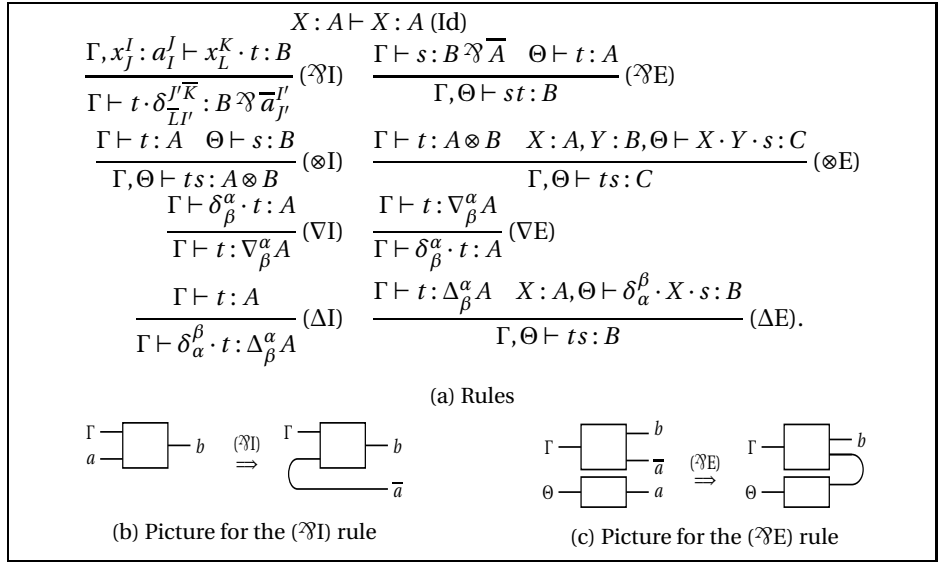

\begin{subfigure}{1\textwidth}
 \centering
$X:A\vdash X:A~(\rm {Id})\quad\quad\quad\quad\quad\quad\quad\quad$\\
$
\begin{array}{rl}
\cfrac{\Gamma,x^I_J:a^J_I\vdash x^K_L\cdot t:B}{\Gamma\vdash t\cdot\delta^{J'\overline{K}}_{\overline{L}I'}:B\parr \overline{a}^{{I'}}_{{J'}}}
~(\parr{\rm{I}})
&
\cfrac{\Gamma\vdash s:B\parr \overline{A}\quad\Theta\vdash t:{A}}{\Gamma,\Theta\vdash st:B}~(\parr{\rm{E}})\\
\cfrac{\Gamma\vdash t:A\quad \Theta \vdash s:B}{\Gamma,\Theta\vdash ts:A\otimes B}~(\otimes{\rm{I}})
&
\cfrac{\Gamma\vdash t:A\otimes B\quad X:A,Y:B,\Theta\vdash X\cdot Y\cdot s:C}{\Gamma,\Theta\vdash ts:C}~ (\otimes{\rm{E}})\\
\cfrac{\Gamma\vdash \delta^\alpha_\beta\cdot t:A}{\Gamma\vdash t:\nabla^\alpha_\beta A}~(\nabla{\rm{I}})
&
\cfrac{\Gamma\vdash t:\nabla^\alpha_\beta A}{\Gamma\vdash \delta^\alpha_\beta\cdot t:A}~(\nabla{\rm{E}})\\
\cfrac{\Gamma\vdash  t:A}{\Gamma\vdash \delta_\alpha^\beta\cdot t:\Delta^\alpha_\beta A}~(\Delta{\rm{I}})
&
\cfrac{\Gamma\vdash   t:\Delta^\alpha_\beta A\quad X:A,\Theta\vdash\delta_\alpha^\beta\cdot X\cdot s:B }{\Gamma,\Theta\vdash ts:B}~ (\Delta{\rm{E}}).
\end{array}$
\caption{Rules}
\label{ND ETTC}
\end{subfigure}
%
\begin{subfigure}{.5\textwidth}
 \centering
 \scalebox{.6}[.8]
 {
  \tikz[scale=.5]
         {
             \begin{scope}[shift={(1,0)}]
                \begin{scope}[shift={(.5,0)}]
                     \draw[thick](0,-2)rectangle(2,-.5);
                        \draw[thick,-](-1,-.8)--(0,-.8);

                     \node[left] at(-1,-.8) {$\Gamma$};
                    \draw[thick,-](-1,-1.7)--(0,-1.7);
                      \node[left] at(-1,-1.7) {$a$};

                     \draw[thick,-](2,-1.25)--(3,-1.25);
                     \node[right] at(3,-1.25) {$b$};
                \end{scope}
                \node[below] at(6.,-1.5) {$\Longrightarrow$};
                \node[above] at(6.,-1.7) {$(\parr{\rm{I}})$};
             \end{scope}

                \begin{scope}[shift={(10.25,0)}]
                     \draw[thick](0,-2)rectangle(2,-.5);
                        \draw[thick,-](-1,-.8)--(0,-.8);

                     \node[left] at(-1,-.8) {$\Gamma$};
                    \draw[thick,-](0,-1.7)--(-1,-1.7)
                    to[out=180,in=90]
                    (-1.4,-2.2)
                    to[out=-90,in=180]
                    (-1,-2.7)
                    --(3,-2.7);
                      \node[right] at(3,-2.7) {$\overline{a}$};

                     \draw[thick,-](2,-1.25)--(3,-1.25);
                     \node[right] at(3,-1.25) {$b$};

                \end{scope}
         }
  }
  \caption{Picture for the $(\parr{\rm{I}})$ rule}
  \label{par intro picture}
 \end{subfigure}
\begin{subfigure}{.5\textwidth}
 \centering
 \scalebox{.6}[.8]
 {
    \tikz[scale=.5]
         {
            \begin{scope}[shift={(18.5,1.8)}]
                \begin{scope}[shift={(-1,-1.75)}]
                    \draw[thick](0,0)rectangle(2,1);
                     \draw[thick,-](2,.5)--(3,.5);

                     \node[right] at(3,.5) {$a$};
                     \draw[thick,-](-1,.5)--(0,.5);

                     \node[left] at(-1,.5) {$\Theta$};
                \end{scope}

                \begin{scope}[shift={(-1,1.5)}]
                        \draw[thick](0,-2)rectangle(2,-.5);
                    \draw[thick,-](2,-.8)--(3,-.8);

                     \node[right] at(3,-1.9) {$\overline{a}$};
                    \draw[thick,-](2,-1.7)--(3,-1.7);
                      \node[right] at(3,-.6) {$b$};

                     \draw[thick,-](-1,-1.25)--(0,-1.25);
                     \node[left] at(-1,-1.25) {$\Gamma$};
                \end{scope}

                \node[below] at(4.5,-.5) {$\Longrightarrow$};
                \node[above] at(4.5,-.8) {$(\parr{\rm{E}})$};
                \begin{scope}[shift={(1,0)}]

                    \draw[thick,-](9,.0)--(10,0)
                         to[out=0,in=90](10.5,-.5)
                         to[out=-90,in=0](10,-1)--(9,-1)
                         ;
                    \begin{scope}[shift={(0,-1.75)}]
                                \draw[thick](7,0)rectangle(9,1);

                         \draw[thick,-](6,.5)--(7,.5);

                         \node[left] at(6,.5) {$\Theta$};
                    \end{scope}
                    \begin{scope}[shift={(0,1.5)}]
                                 \draw[thick](7,-2)rectangle(9,-.5);
                        \draw[thick,-](9,-1)--(10 ,-1);

                        \draw[thick,-](9,-1.5)--(10,-1.5);
                          \node[right] at(10,-1.) {$b$};

                         \draw[thick,-](6,-1.25)--(7,-1.25);
                         \node[left] at(6,-1.25) {$\Gamma$};
                    \end{scope}

            \end{scope}

         \end{scope}

         }
    }
 \caption{Picture for the $(\parr{\rm{E}})$ rule}
  \label{par elim picture}
\end{subfigure}
\caption{Natural deduction for {\bf ETTC} }
\end{figure}

An {\it n. d. tensor typing judgement} is an expression of the form $t:A$, where $t$ is a natural deduction term and $A$ is a tensor type, with $Fsup(t)=Fsub(A)$, $Fsub(t)=Fsup(A)$ (basically, we use a colon instead of a turnstile).
A {\it variable declaration} is a typing judgement with a tensor variable to the left of the colon, and an {\it n.d. typing context} is a finite set of variable declarations which have no common variable symbols. An  {\it n.d. tensor sequent} is an expression of the form $\Gamma\vdash \sigma$, where $\Gamma$ is an n.d. typing context and $\sigma$ is an n.d. typing judgement. {\it Natural deduction system of} {\bf ETTC} is given by the rules in Figure \ref{ND ETTC}.

Again, it is implicit that types and typing judgements in Figure \ref{ND ETTC}
are well defined. In particular, in the $(\parr{\rm{I}})$ rule we need pairwise disjoint sequences $I',J'$ of pairwise distinct fresh indices, in the $(\parr{\rm{E}})$ rule the contexts $\Gamma,\Theta$ have no common tensor variable symbols etc.
The $(\nabla{\rm{I}})$ and $(\Delta{\rm{I}})$ rules are same as in the sequent calculus and use the same convention: $\alpha\in Fsup(A)$, $\beta\in Fsub(A)$.

 Let us say that a n.d. term is {\it closed} if it contains no tensor variables. A closed term is {\it pure} if it contains no terminal symbols. A closed term is {\it lexical} if, on the contrary, it cannot be written as a normal term expression of the form $\delta^i_j\cdot t'$ (in a geometric language, all edges of the corresponding graph have nonempty labels). A typing judgement $t:A$ is {\it lexicalized} if the term $t$ is lexical.

It is easily seen that any sequent  derivable in natural deduction of {\bf ETTC} can be written in the form
\be\label{ND derivable sequent}
X_{(1)}:(a_{(1)})^{I_1}_{J_1},\ldots,X_{(n)}:(a_{(n)})^{I_n}_{J_n}\vdash X_{(1)}\cdots X_{(n)}\cdot t:B,
\ee
where the term $t$ is pure. (In the case of the $({\rm{Id}})$ axiom we  put $t=1$.)
The following is completely standard.
\bl[``Deduction theorem'']\label{extended deduction theorem}
Let $\Xi$ be  a finite set of lexicalized typing judgements,
$\Xi=\{
\tau_{(1)}:(a_{(1)})^{I_1}_{J_1},\ldots,\tau_{(n)}:(a_{(n)})^{I_n}_{J_n}\}$.

A natural deduction sequent of the form $\vdash\sigma:B$
 is derivable
 from  $\Xi$ using each element of $\Xi$ exactly once  iff there is a natural deduction sequent of the form (\ref{ND derivable sequent}) derivable without nonlogical axioms such that $t\cdot\tau_{(1)}\ldots\tau_{(n)}=\sigma$. $\Box$
\el

  We define the {\it sequent calculus translation} of (\ref{ND derivable sequent}) as the tensor typing judgement
\be\label{ND translation}
\delta^{I_n'\overline{J_n}}_{\overline{I_n}{J_n'}}\cdots \delta^{I_1'\overline{J_1}}_{\overline{I_1}{J_1'}}\cdot t\vdash B, \overline{(a_{(n)})}^{J_n'}_{I_n'},\ldots,\overline{(a_{(1)})}^{J_1'}_{I_1'},
\ee
where $I_1',\ldots,I_n'$, $J_1',\ldots,J_n'$ are pairwise disjoint sequences of pairwise distinct indices,  all disjoint from ${FSup}(B)$, ${FSup}(B)$.
Translation (\ref{ND translation}) provides for equivalence of natural deduction and sequent calculus formulation of {\bf ETTC}, which is proven in a routine way familiar from the case of, say, {\bf MILL1}.
\bl\label{ND=sequent calculus}
Sequent (\ref{ND derivable sequent}) is derivable in the natural deduction formulation of {\bf ETTC} iff its translation (\ref{ND translation}) is derivable in the sequent calculus formulation. $\Box$
\el

\subsection{Geometric representation and example}
Translating to the sequent calculus allows geometric representation of natural deduction derivations as operations on bipartite graphs with indices corresponding to vertices. It is natural to depict indices occurring to the left and to the right of the turnstile as vertices aligned on two parallel lines (continuously deforming the graph of translation (\ref{ND translation}), where all vertices are on one line). Also, it should be clear from  formula (\ref{ND translation}) that  different occurrences of the same index  in a natural deduction sequent correspond to distinct indices in the translation and to distinct vertices in the picture. In Figures \ref{par intro picture}, \ref{par elim picture}, \ref{ND picture} we align vertices vertically. With such a convention, the geometric representation  for the $(\parr)$ rules can be shown schematically as in Figures \ref{par intro picture}, \ref{par elim picture}. The $(\parr{\rm{I}})$ rule does not change the graph, but changes its pictorial representation by a continuous deformation.

Figure \ref{ND derivation} shows a derivation of a tensor sequent corresponding under the translation in Figure \ref{translating lambek types} to a  sequent derivable in {\bf LC}. Figure \ref{ND picture} gives a geometric representation of the derivation.

\begin{figure}[t]
\begin{subfigure}{1\textwidth}
\centering
\begin{prooftree}
\def\fCenter{\ \vdash\ }
\AxiomC{$y^{j}_m:\nabla^k_i(b^m_k\parr\overline{a}^i_j) \vdash y^{j}_m:\nabla^k_i(b^m_k\parr\overline{a}^i_j)$}
\RightLabel{$(\nabla{\rm {E}})$}
\UnaryInf$y^{j}_m:\nabla^k_i(b^m_k\parr\overline{a}^i_j)  \fCenter \delta^k_i y^{j}_m:b^m_k\parr\overline{a}^i_j $
\RightLabel{$(=)$}
\UnaryInf$ y^{j}_m:(b/a)^m_j \fCenter \delta^k_i y^{j}_m:b^m_k\parr\overline{a}^i_j$
\AxiomC{$x^{i}_j:a^j_i \vdash x^{i}_j:a^j_i$}
\RightLabel{$(\parr{\rm {E}})$}
\BinaryInf$ y^{j}_m:(b/a)^m_j,x^{u}_v:a^v_u \fCenter \delta^k_i y^{j}_mx^{i}_j:b^m_k$
\RightLabel{$(=)$}
\UnaryInf$ y^{j}_m:(b/a)^m_j,x^{u}_v:a^v_u \fCenter  y^{j}_m x^{k}_j:b^m_k$
\end{prooftree}
\caption{Derivation}
\label{ND derivation}
\end{subfigure}
\begin{subfigure}{1\textwidth}
\centering
\tikz[scale=.5]
{
\begin{scope}[shift={(0,0)}]
    \begin{scope}[shift={(-2.5,0)}]

        \begin{scope}[shift={(0,-4)}]
             \draw[thick,<-](1,2)--(2,2);
             \draw [fill] (2,2) circle [radius=0.05];
             \draw[thick,->](1,1.3)--(2,1.3);
             \draw [fill] (1,1.3) circle [radius=0.05];
            \node[left]at (1,2){$j$};
            \node[right]at (2,2){$j$};
            \node[left]at (1,1.3){$i$};
            \node[right]at (2,1.3){$i$};
            \draw[dashed,](.5,2.2)--(0.2,2.2)--(0.2,1.2)--(.5,1.2);
            \node[left] at (.2,1.7){$a^j_i$};
            \draw[dashed,](2.5,2.2)--(2.8,2.2)--(2.8,1.2)--(2.5,1.2);
            \node[right] at (2.8,1.7){$a^j_i$};
        \end{scope}

        \begin{scope}
                \draw[thick,<-](1,.4)--(2,.4);
              \draw[thick,->](1,-.3)--(2,-.3);
              \draw [fill] (2,.4) circle [radius=0.05];
              \draw [fill] (1,-.3) circle [radius=0.05];
              \node[left]at (1,.4){$m$};
            \node[right]at (2,.4){$m$};
            \node[left]at (1,-.3){$j$};
            \node[right]at (2,-.3){$j$};            \draw[dashed,](.2,.6)--(-0.1,.6)--(-0.1,-.5)--(.2,-.5);
            \node[left] at (0.,.05){$\nabla^k_i(b^m_k\parr\overline{a}_i^j)$};
            \draw[dashed,](2.8,.6)--(3.1,.6)--(3.1,-.5)--(2.8,-.5);
            \node[right] at (3,.05){$\nabla^k_i(b^m_k\parr\overline{a}_i^j)$};
        \end{scope}

         \draw(-4.25,1.5)rectangle(7.25,-3.4);

    \end{scope}
    \node [below]at(6.25,-1.05) {$\Longrightarrow$};
    \node [above]at(6.25,-1.35) {$(\nabla{\rm {E}})$};
\end{scope}

\begin{scope}[shift={(14,0)}]
    \begin{scope}[shift={(-1.5,0)}]

        \begin{scope}[shift={(0,-4)}]
             \draw[thick,<-](1,2)--(2,2);
             \draw [fill] (2,2) circle [radius=0.05];
             \draw[thick,->](1,1.3)--(2,1.3);
             \draw [fill] (1,1.3) circle [radius=0.05];
            \node[left]at (1,2){$j$};
            \node[right]at (2,2){$j$};
            \node[left]at (1,1.3){$i$};
            \node[right]at (2,1.3){$i$};
            \draw[dashed,](.5,2.2)--(0.2,2.2)--(0.2,1.2)--(.5,1.2);
            \node[left] at (.2,1.7){$a^j_i$};
            \draw[dashed,](2.5,2.2)--(2.8,2.2)--(2.8,1.2)--(2.5,1.2);
            \node[right] at (2.8,1.7){$a^j_i$};
        \end{scope}

        \begin{scope}
                \draw[thick,<-](1,.4)
                to[out=0,in=-90](1.5,.7)
                to[out=90,in=0](2,1);
                \draw [fill] (2,1) circle [radius=0.05];

              \draw[thick,->](1,-.3)
              to[out=0,in=90](1.5,-.6)
                to[out=-90,in=180](2,-.9);
                \draw [fill] (1,-.3) circle [radius=0.05];
              \node[left]at (1,.4){$m$};
            \node[right]at (2,1){$m$};

            \node[right]at (2,.4){$k$};

            \draw[thick,->](2,-.3)
              to[out=180,in=-90](1.5,.05)
                to[out=90,in=180](2,.4);
                \draw [fill] (2,-3) circle [radius=0.05];

            \node[right]at (2,-.3){$i$};
            \node[left]at (1,-.3){$j$};
            \node[right]at (2,-.9){$j$};
            \draw[dashed,](.2,.6)--(-0.1,.6)--(-0.1,-.5)--(.2,-.5);
            \node[left] at (0.,.05){$\nabla^k_i(b^m_k\parr\overline{a}^i_j)$};
            \draw[dashed,](2.8,1.2)--(3.1,1.2)--(3.1,.2)--(2.8,.2);
            \draw[dashed,](2.6,-.2)--(3.1,-.2)--(3.1,-1.)--(2.6,-1.);
            \node[right] at (3,.7){$b^m_k$};
            \node[right] at (3,-.6){$\overline{a}^i_j$};
            \draw(-4.4,1.5)rectangle(4.25,-3.4);
        \end{scope}

    \end{scope}
\end{scope}

\begin{scope}[shift={(-2.5,-5.35)}]
    \begin{scope}
        \node at(-2.2,-.9) {$=$};
            \begin{scope}[shift={(0,-4)}]
                         \draw[thick,<-](1,2)--(2,2);
                         \draw[thick,->](1,1.3)--(2,1.3);
                         \draw [fill] (2,2) circle [radius=0.05];
                         \draw [fill] (1,1.3) circle [radius=0.05];

                        \node[left] at (1,1.7){$a$};
                        \node[right] at (2,1.7){$a$};
            \end{scope}

            \begin{scope}
                    \draw[thick,<-](1,.4)
                    to[out=0,in=-90](1.5,.7)
                    to[out=90,in=0](2,1);
                    \draw [fill] (2,1) circle [radius=0.05];

                  \draw[thick,->](1,-.3)
                  to[out=0,in=90](1.5,-.6)
                    to[out=-90,in=180](2,-.9);
                    \draw [fill] (1,-.3) circle [radius=0.05];


                \draw[thick,->](2,-.3)
                  to[out=180,in=-90](1.5,.05)
                    to[out=90,in=180](2,.4);

                \node[left] at (1.,.05){$b/a$};
                \node[right] at (2,.7){$b$};
                \node[right] at (2,-.6){$\overline{a}$};
                \draw(-.6,1.5)rectangle(3.05,-3.4);
            \end{scope}

    \end{scope}
\end{scope}

\begin{scope}[shift={(-2.5,0)}]

     \begin{scope}[shift={(6.9,-5.35)}]
         \node [below]at(-2.1,-.9) {$\Longrightarrow$};
         \node [above]at(-2.1,-1.2) {$(\parr{\rm {E}})$};
         \begin{scope}

                 \begin{scope}[shift={(0,-4)}]
                     \draw[thick,<-](1,2)--(2,2);
                     \draw[thick,-](1,1.3)--(2,1.3);
                     \draw [fill] (1,1.3) circle [radius=0.05];
                    \node[left] at (1,1.7){$a$};

                \end{scope}

                \begin{scope}
                        \draw[thick,<-](1,.4)
                        to[out=0,in=-90](1.5,.7)
                        to[out=90,in=0](2,1);
                        \draw [fill] (2,1) circle [radius=0.05];

                      \draw[thick,-](1,-.3)
                      to[out=0,in=90](1.5,-.6)
                        to[out=-90,in=180](2,-.9)
                           to[out=0,in=90](2.3,-1.45)
                           to[out=-90,in=0](2,-2);
                           \draw [fill] (1,-.3) circle [radius=0.05];

        %

                    \draw[thick,->]
                    (2,-2.7)
                    to[out=0,in=-90](2.7,-1.45)
                    to[out=90,in=0]
                    (2,-.3)
                      to[out=180,in=-90](1.5,.05)
                        to[out=90,in=180](2,.4);

                    \node[left] at (1,.05){$b/a$};
                    \node[right] at (2,.7){$b$};
                    \draw(-.6,1.5)rectangle(3.05,-3.4);
                \end{scope}
            \end{scope}

            \begin{scope}[shift={(8,0)}]
                \node at(-2.7,-.9) {$=$};
                \begin{scope}

                    \begin{scope}[shift={(0,-4)}]
                         \draw[thick,<-](1,2)--(2,2);
                         \draw[thick,-](1,1.3)--(2,1.3);
                         \draw [fill] (1,1.3) circle [radius=0.05];

                        \node[left] at (1,1.7){$a$};
                    \end{scope}

                    \begin{scope}
                            \draw[thick,<-](1,.4)
                                                 to[out=0,in=90](2.5,-.25)
                            to[out=-90,in=180](3,-.9);
                            \draw [fill] (3,.-.9) circle [radius=0.05];

                          \draw[thick,-](1,-.3)
                               to[out=0,in=90](2.25,-1.15)
                               to[out=-90,in=0](2,-2);
                          \draw [fill] (1,-.3) circle [radius=0.05];

                        \draw[thick,->]
                        (2,-2.7)
                         to[out=0,in=-90](2.5,-1.55)
                            to[out=90,in=180](3,-1.4);

                        \node[left] at (1,.05){$b/a$};

                        \node[right] at (3,-1.15){$b$};
                        \draw(-.6,1.5)rectangle(3.9,-3.4);

                    \end{scope}
                \end{scope}
            \end{scope}
        \end{scope}
     \end{scope}

}
\caption{Geometric representation}
\label{ND picture}
\end{subfigure}
\caption{Example for natural deduction}
\label{ND example}
\end{figure}
\subsection{Grammars}\label{tensor grammars section}
\begin{figure}
\begin{subfigure}{\textwidth}
\centering
$
\cfrac{\Gamma,x^I_J:a^J_I\vdash x^K_L\cdot t:B}
{\Gamma\vdash t\cdot\delta^{\overline{K}J'}_{I'\overline{L}}:{a}^{{I'}}_{{J'}}\multimap B }
~(\multimap{\rm{I}})\quad
\cfrac{\Gamma\vdash s:A\multimap B\quad\Theta\vdash t:{A}}{\Gamma,\Theta\vdash st:B}~(\multimap{\rm{E}})$
\\
$
\cfrac
{
    \Gamma\vdash t:(b/a)^i_j\quad\Theta\vdash s:a^j_k
}
{
\Gamma,\Theta\vdash t s:b^i_k
}
(/ {\rm {E}})
\quad
\cfrac
{
\Gamma\vdash s:a^j_k\quad\Theta\vdash t:(a\backslash b)^i_j
}
{
\Gamma,\Theta\vdash s t:b^i_k
}
(\backslash{\rm{E}})
$
\caption{Admissible rules}
\label{Admissible rules}
\end{subfigure}
\begin{subfigure}{\textwidth}
\centering
$$[{\rm{Mary}}]^i_j: np,~
[{\rm{John}}]^i_j: np,~
[{\rm{loves}}]^i_j:(np\backslash s)/np,
$$
$$[{\rm{madly}}]^i_j:((np\backslash s)\backslash(np\backslash )),~
[{\rm{who}}]^i_j:
(np\backslash np)/\Delta^u_t(np^t_u\multimap s).$$
\caption{Axioms}
\label{Axioms}
\end{subfigure}
\begin{subfigure}{\textwidth}
\centering
$\mbox{ }$\\
   \begin{prooftree}
    \def\fCenter{\mbox{\ $\vdash$\ }}
    \def\ScoreOverhang{.1pt}
    \Axiom$
            \fCenter [{\rm{loves}}]^k_l:(np\backslash s)/np
                $
     \Axiom$x^j_k:np\fCenter x^j_k:np$
    \RightLabel{$(/{\rm {E}})$}
    \insertBetweenHyps{\hskip 5pt}
    \BinaryInf$
        x^j_k:np \fCenter
          [{\rm{loves}}]^k_l x^j_k:np\backslash s
        $
     \Axiom$
            \fCenter [{\rm{madly}}]^i_j:
            (np\backslash s)\backslash(np\backslash s)
            $
    \RightLabel{$(\backslash{\rm {E}})$
    \insertBetweenHyps{\hskip 5pt}}
    \BinaryInf$
        x^j_k:np \fCenter
        [{\rm{loves}}]^k_l x^j_k[{\rm{madly}}]^i_j:
        np\backslash s
        $
    \end{prooftree}

    \begin{prooftree}
    \def\fCenter{\mbox{\ $\vdash$\ }}
    \def\ScoreOverhang{.1pt}
    \Axiom$
            \fCenter [{\rm{John}}]^l_m:
            np
            $
    \Axiom$
            x^j_k:np \fCenter
        [{\rm{loves}}]^k_l x^j_k[{\rm{madly}}]^i_j:
        np\backslash s
                $
     \RightLabel{$(\backslash{\rm {E}})$}
    \insertBetweenHyps{\hskip 5pt}
    \BinaryInf$
        x^j_k:np \fCenter
          [{\rm{John}}\mbox{ }{\rm{loves}}]^k_m x^j_k[{\rm{madly}}]^i_j: s_i^m
        $
    \RightLabel{$(\multimap{\rm {I}})$}
    \UnaryInf$
        \fCenter
            [{\rm{John}}\mbox{ }{\rm{loves}}]^k_m x^j_k[{\rm{madly}}]^i_j \delta^{j\alpha}_{\beta k}:
        np^\beta_\alpha\multimap s_i^m
        $
    \RightLabel{$(=)$}
    \UnaryInf$
    \fCenter
          [{\rm{John}}\mbox{ }{\rm{loves}}]^\alpha_m \cdot[{\rm{madly}}]^i_\beta:
        np^\beta_\alpha\multimap s_i^m
        $
    \RightLabel{$(\Delta{\rm {I}})$}
    \UnaryInf$
    \fCenter
          \delta_\alpha^\beta\cdot[{\rm{John}}\mbox{ }{\rm{loves}}]^\alpha_m \cdot[{\rm{madly}}]^i_\beta:
        \Delta^\alpha_\beta(np^\beta_\alpha\multimap s_i^m)
        $
        \RightLabel{$(=)$}
    \UnaryInf$
    \fCenter
          [{\rm{John}}\mbox{ }{\rm{loves}}\mbox{ }{\rm{madly}}]^i_m:
        \Delta^\alpha_\beta(np^\beta_\alpha\multimap s)
        $
    \end{prooftree}

\begin{prooftree}
    \def\fCenter{\mbox{\ $\vdash$\ }}
    \def\ScoreOverhang{.1pt}
    \Axiom$
            \fCenter [{\rm{who}}]_i^j:
    (np\backslash np)/\Delta^\beta_\alpha(np^\alpha_\beta\multimap s)
            $
    \Axiom$
             \fCenter
        [{\rm{John}}~{\rm{loves}}~{\rm{madly}}]^i_n:
        \Delta^\alpha_\beta(np^\beta_\alpha\multimap s)
                $
    \insertBetweenHyps{\hskip 5pt}
    \RightLabel{$(/{\rm {E}})$}
    \BinaryInf$
         \fCenter
          [{\rm{who}}\mbox{ }{\rm{John}}\mbox{ }{\rm{loves}}\mbox{ }{\rm{madly}}]^j_n:
        np\backslash np
        $
    \end{prooftree}

\begin{prooftree}
    \def\fCenter{\mbox{\ $\vdash$\ }}
    \def\ScoreOverhang{.1pt}
    \Axiom$
            \fCenter [{\rm{Mary}}]^i_j:
            np
            $
    \Axiom$
         \fCenter
          [{\rm{who}}\mbox{ }{\rm{John}}\mbox{ }{\rm{loves}}\mbox{ }{\rm{madly}}]^j_n:
        np\backslash np
        $
    \RightLabel{$(\backslash{\rm {E}})$}
    \insertBetweenHyps{\hskip 5pt}
    \BinaryInf$
        \fCenter
          [{\rm{Mary}}\mbox{ }{\rm{who}}\mbox{ }{\rm{John}}\mbox{ }{\rm{loves}}\mbox{ }{\rm{madly}}]^i_n:
         np
        $
    \end{prooftree}
\caption{Derivation}
\label{ND grammar derivation}
\end{subfigure}
\caption{Example with nonlogical axioms}
\label{ND grammar example}
\end{figure}
We give a more elaborate, linguistically motivated example of a derivation from nonlogical axioms in Figure \ref{ND grammar example}. For better readability, we systematically use notation of {\bf LC} and {\bf MILL} understood as in Figure \ref{translating lambek types} and, in concrete derivations, we omit free indices in types of valency $(1,1)$  (they are uniquely determined by free indices in terms).  In Figure \ref{Admissible rules} we  fix  admissible rules corresponding to rules of {\bf LC} and {\bf MILL}  that we will use. Figure \ref{Axioms} shows our axioms (the lexicon), which we assume closed under $\alpha$-equivalence of typing judgements, and in Figure \ref{ND grammar derivation} we derive the noun phrase ``Mary who John loves madly''. It might be entertaining to reproduce the derivation in the geometric language.

Now let us define a {\it tensor lexical entry} as an $\alpha$-equivalence class of  lexicalized typing judgements and a {\it tensor grammar} as a pair $(Lex,s)$, where $Lex$ is a finite set of lexical entries and $s$ is an atomic type symbol of valency $(1,1)$. For a tensor grammar $G$ we will right $\vdash_G \sigma:A$ to indicate that the sequent $\sigma:A$ is derivable in {\bf ETTC} from elements of $Lex$.
We define {\it two} languages generated by $G$. The {\it regular language $L(G)$} and the {\it regularized language $L_{reg}(G)$} of $G$ are, respectively   the sets
$$L(G)=\{w|~\vdash_G [w]^i_j:s\},\quad L_{reg}(G)=\{w|~\vdash_G [w]^i_j\cdot(\delta)^n:s\}.$$
(Note that $n$ in the definition of $L_{reg}$ is not an index, but indicates the $n$-th power.)

Any balanced {\bf MILL1} grammar $G$  translates to a tensor grammar $\widetilde G$ by writing for every lexical entry $(w,A)$ of $G$ the tensor lexical entry $[w]_{\rho(r)}^{\pi(l)}:a^{\rho(r)}_{\pi(l)}$, where
$a$ is the type symbol of $||A||^\rho_\pi$ (the choice of particular bijections  $\rho$, $\pi$ is not important because of closure under $\alpha$-equivalence of typing judgements).
\bl In the above setting we have $L(G)=L_{reg}(\widetilde G)$.
\el
{\bf Proof} Let $w\in L_{reg}(\widetilde G)$. Then we have $\vdash_G \tau:s$, where $\tau=[w]^i_j\cdot(\delta)^n$.

Obviously $w$ is the concatenation $w=w_1\ldots w_n$ of all words occurring in lexical entries $[w_\mu]^i_j:(a_{(\mu)})_i^j$, $\mu=1,\ldots,n$,
used to derive $\vdash \tau:S$. By Lemma \ref{extended deduction theorem} (``Deduction theorem'')  it must be that
the we have the derivable n.d. sequent
$$
(x_{(1)})^{i_1}_{j_1}:(a_{(1)})_{i_1}^{j_1},\ldots,(x_{(n)})^{i_n}_{j_n}:(a_{(n)})_{i_n}^{j_n}\vdash    \delta_{i_1}^{j_2}\cdots\delta_{i_{n-1}}^{j_{n}}\cdot(\delta)^n\cdot (x_{(1)})^{i_1}_{j_1}\cdots (x_{(n)})^{i_n}_{j_n}:s_{i_n}^{j_1},
$$
which corresponds, by Lemma \ref{ND=sequent calculus}, to the derivable typing judgement
\be\label{grammar_deduction}
\delta_{i_0}^{j_1}\cdots\delta_{i_n}^{j_{n+1}}\cdot(\delta)^n\vdash s^{i_0}_{j_{n+1}},\overline{(a_{(n)})}_{j_{n}}^{i_n},\ldots,\overline{(a_{(1)}})_{j_1}^{i_1}.
\ee
If we translate  {\bf MLL1} to {\bf ETTC} using the prescription  $\rho(\mu)=i_\mu$, $\pi(\mu)=i_{\mu+1}$, where $\mu=0,\ldots,n$, then, by  Lemma \ref{FO2ETTC}, we have that (\ref{grammar_deduction}) comes from a derivation of the sequent
$\vdash S[0;n],\overline{A_n}[n;n-1],\ldots, \overline{A_1}[1;0]$.
 The latter is identified as the image of an {\bf MILL1} sequent expressing, by definition, that $w\in L(G)$.

The opposite inclusion is similar, just easier. $\Box$

  Geometric intuition  might suggest that it is the regular, rather than, regularized, language of a tensor grammar that should be accepted as natural.
In any case, it can be shown directly that in the case of   tensor grammars obtained from Lambek grammars or $\lambda$-grammars there are no additional words generated ``by regularization'': i.e. $L(G)=L_{reg}(G)$.

 \end{document}